\documentclass{article}

\usepackage{microtype}
\usepackage{graphicx}
\usepackage{booktabs}
\usepackage{multirow}
\usepackage{amsmath}
\usepackage{amssymb}
\usepackage{xcolor}
\usepackage{colortbl}
\usepackage{array}
\usepackage{hyperref}

\usepackage[accepted]{icml2026}
\makeatletter
\renewcommand{\ICML@appearing}{\textit{Accepted at the ICML 2026 Workshop on Structured Data for Health (SD4H).}}
\makeatother
\usepackage[capitalize,noabbrev]{cleveref}

\newcommand{\metric}[1]{\textsc{#1}}
\newcommand{\condition}[1]{\textsc{#1}}
\newcommand{\systemname}{\textsc{TFTS}}

\icmltitlerunning{Partitioning Deterministic and Neural Health Generation}

\begin{document}

\twocolumn[
  \icmltitle{Think Fast, Talk Smart:
  Partitioning Deterministic and Neural Computation for Structured Health Text Generation}

  \begin{icmlauthorlist}
  \icmlauthor{Kai-Chen Cheng}{eightsleep}
  \icmlauthor{Haejun Han}{eightsleep}
  \icmlauthor{David Q. Sun}{eightsleep}
    \end{icmlauthorlist}
    \icmlaffiliation{eightsleep}{AI/ML@Eight Sleep}
    \icmlcorrespondingauthor{Kai-Chen Cheng}{kevin.cheng@eightsleep.com}
  \icmlkeywords{Structured Health Data, Large Language Models, Hybrid Systems, Deterministic Evaluation, Health Insight Generation}
  \vskip 0.3in
]

\printAffiliationsAndNotice{}

\begin{abstract}
Large language models (LLMs) are increasingly being used to generate health text from structured records such as wearable time series, biomarkers, vitals, and care-management logs.
For recurring health outputs, fluency is not enough: systems must remain faithful to source data, ground explanatory claims in available evidence, follow stated policies, emit machine-readable outputs, and run cheaply enough for repeated use.
We ask which responsibilities in structured health generation should be deterministic computation rather than runtime LLM prompting.
We introduce Think Fast, Talk Smart (\systemname{}), a sleep-health insight pipeline in which deterministic code performs recurring analysis before one bounded LLM writer call.
Across 280 user-nights and six models, \systemname{} achieves lower numeric error, lower instruction-compliance error, and lower end-to-end cost than structured zero-shot and few-shot one-call baselines.
Layer replacement reveals contract-specific failures: LLM comparison raises numeric error, LLM ranking degrades policy selection, LLM attribution increases unsupported causal language, and an LLM-generated writer interface reintroduces errors even after upstream facts are deterministic.
The results support a broader design rule: let code own recurring analysis, and let LLMs express verified facts within bounded interfaces.
\end{abstract}

\section{Introduction}
\label{sec:intro}

Many health and medical applications ask LLMs to turn structured records into user-facing text: wearable coaching summaries, lab results interpretations, medication guidance, vitals monitoring, and care-management reports \citep{thirunavukarasu2023large,singhal2023towards}.
These applications are promising precisely because language models can make technical records understandable. They are also risky because fluent health text must remain faithful to the source record, consistent with task instructions, and explicit about what evidence supports each claim \citep{ji2023survey,vanveen2024adapted}.
Recent wearable and personal-health systems show the promise of LLMs in this space \citep{kim2024healthllm,khasentino2025phllm,merrill2026phia}; they also sharpen a practical question: which parts of a recurring health-generation task should be learned generation, and which should be verified computation?

Prompting research offers several ways to improve model reasoning.
Chain-of-thought elicits intermediate reasoning steps \citep{wei2022chain}; iterative refinement trades more calls for improved outputs \citep{madaan2023selfrefine}; program-aided methods externalize computation \citep{chen2023program,gao2023pal}; retrieval and tool-use systems augment LLMs at runtime \citep{lewis2020retrieval,yao2023react}; and prompt-programming frameworks such as DSPy optimize LLM pipelines \citep{khattab2023dspy}.
These methods improve what happens inside or around an LLM call. Our work asks a different systems question: when should a recurring health task be implemented outside the LLM call?
For structured health workloads, some decisions recur with little change across inputs, such as comparing a measurement to a baseline, choosing what to surface, and deciding which explanatory claims are supported by logged evidence.
When such steps are stable and verifiable, runtime prompting may be the wrong abstraction for the analytical work.

\begin{figure*}[t]
\centering
\includegraphics[width=0.94\textwidth]{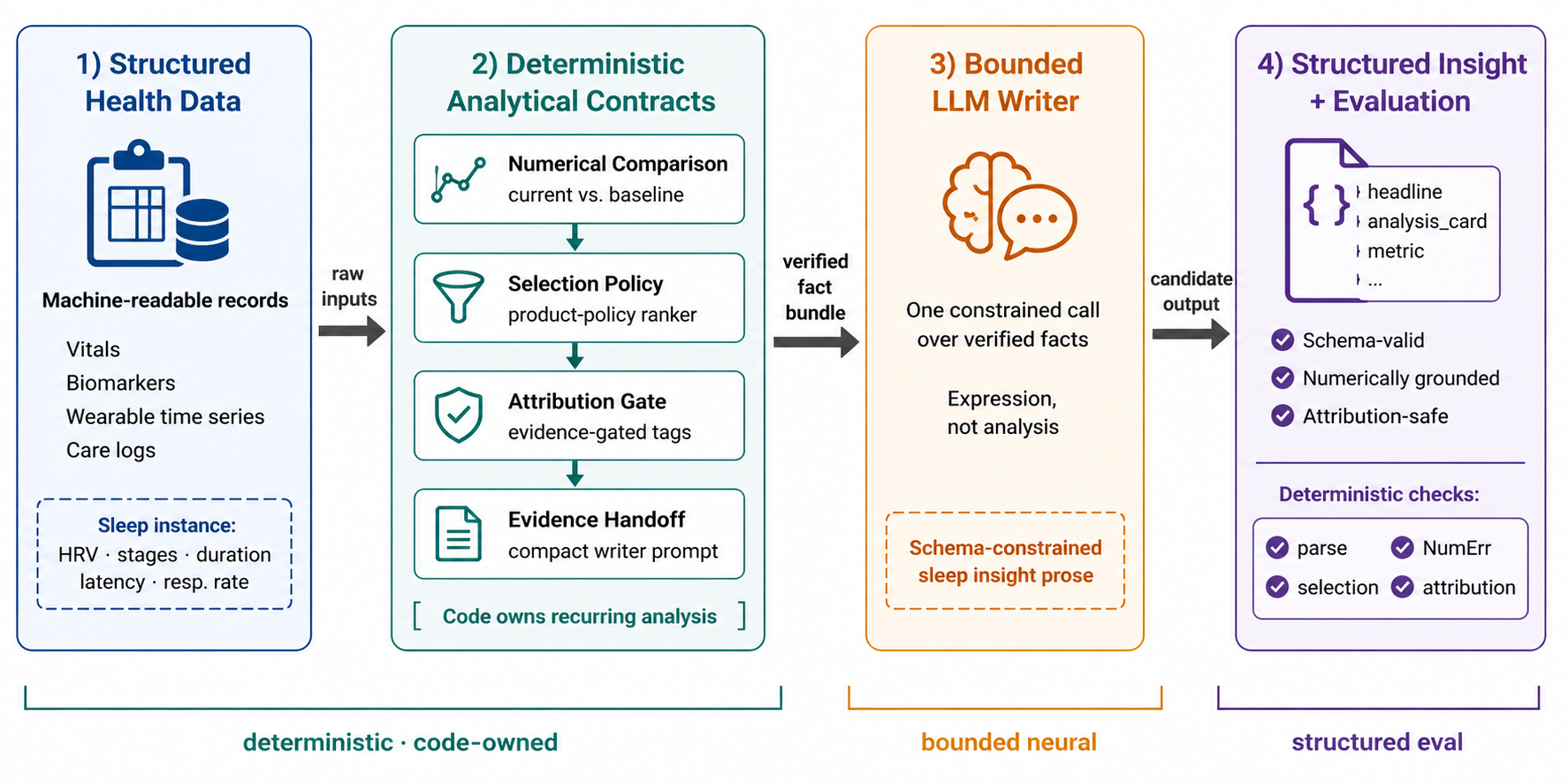}
\caption{\textbf{The \systemname{} partitioning framework.} Recurring structured health-generation tasks are split into structured inputs, deterministic analytical modules, a bounded LLM writer, and deterministic evaluation of the structured output. The sleep-health callouts instantiate this partition in the evaluated pipeline.}
\label{fig:framework}
\end{figure*}

We study this partitioning problem through Think Fast, Talk Smart (\systemname{}), a sleep-health insight pipeline.
Sleep-health is a useful case study because wearable records are longitudinal and heterogeneous, while the final output must be brief, grounded, and usable by non-expert readers.
\systemname{} implements stable analytical steps as deterministic layers and uses an LLM only as a bounded final writer over a compact evidence interface.
We compare this design against one-call baselines, then replace one analytical layer at a time with an LLM-generated typed artifact.
This layer-replacement protocol tests whether a model preserves the intended meaning of an intermediate analytical artifact beyond producing syntactically valid JSON.

Our contributions are threefold: we show that deterministic analytical layers plus a bounded writer achieve lower error and lower cost than strong prompting baselines on recurring health-insight generation tasks; introduce a layer-replacement protocol that isolates individual analytical responsibilities with typed LLM-generated intermediates; and identify the writer interface as a non-obvious failure point.
Although our experiments use sleep-health data, the partitioning principle targets broader structured health settings where inputs are machine-readable, analytical rules recur, and intermediates can be checked.

\section{Task and Pipeline}
\label{sec:task}

\Cref{fig:framework} shows the four-stage partitioning pattern and the sleep-health instance we evaluate.
For each user-night, the system ingests wearable-derived sleep, recovery, physiology, and behavior signals and returns a schema-constrained insight with a title, core insight, improvement suggestion, selected metric metadata, attribution tags, and chart data.
The deterministic reference pipeline follows a classic data-to-text decomposition \citep{reiter2007architecture,wiseman2017challenges}: it formats data, computes numerical comparisons, ranks which metric should be surfaced, gates attribution by evidence, and compacts the selected facts into a deterministic prompt interface.
The final LLM call then produces user-facing prose under a bounded schema, after which deterministic assembly normalizes and wraps the response.

\systemname{} decouples analytical reasoning from linguistic expression: deterministic code handles recurring, verifiable reasoning, while the LLM focuses on phrasing verified facts for the user.
The experiment asks which recurring responsibilities can safely be moved from deterministic code into LLM-generated intermediates.

\section{Baselines and Layer Replacement}
\label{sec:design}

\paragraph{One-call prompting baselines.}
We compare \systemname{} to two single-call alternatives.
\condition{Structured Zero-Shot} receives raw data, the output schema, and the same sleep insight generation rules in prompt form; it must analyze, select, attribute, and write in one call without demonstrations.
\condition{Structured Few-Shot} strengthens this baseline with structured metric tables, few-shot examples, explicit selection rules, and numeric-grounding instructions.

\begin{figure*}[t]
\centering
\includegraphics[width=0.94\textwidth]{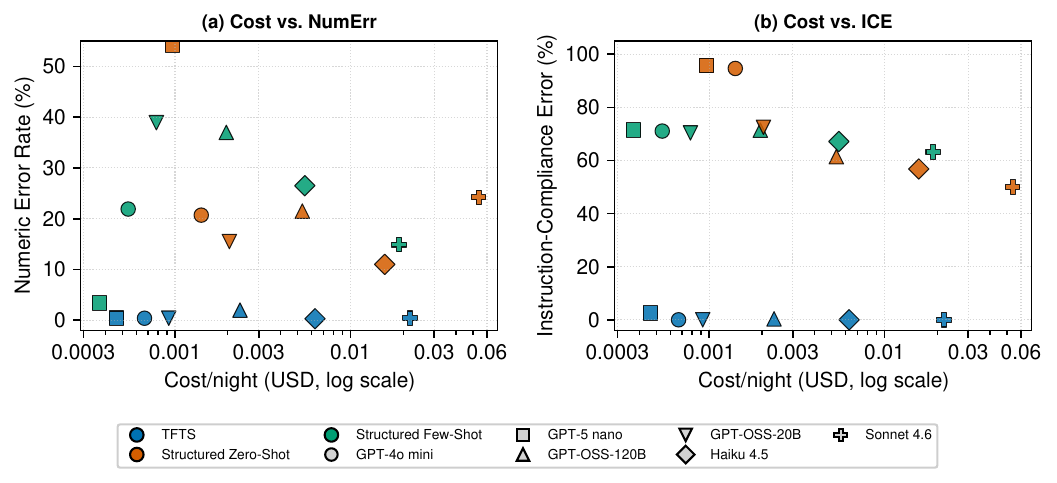}
\caption{Baseline cost/error frontier across six models ($n=280$ per cell). Both panels use end-to-end \metric{Cost/night} on a log scale. \metric{Instruction-Compliance Error} is the union of \metric{SelErr} and \metric{AttrErr}. \systemname{} remains lower on both error axes than the structured zero-shot and few-shot one-call baselines across the evaluated cost range.}
\label{fig:frontier}
\end{figure*}

\paragraph{Layer-replacement protocol.}
For mechanistic analysis, we replace exactly one analytical layer with an LLM-generated typed artifact.
All upstream reference layers remain deterministic; the generated artifact replaces the target layer's reference output.
This protocol is stricter and fairer than removing a layer from the final writer prompt: the model is asked to perform one bounded intermediate task, not to recover missing context while writing the final JSON.
The main conditions replaced \condition{Comparison}, \condition{Ranker}, \condition{Attribution}, and \condition{Handoff} layers.

\paragraph{Data and models.}
The experiment uses 280 user-nights from 20 active users.
We evaluate six models: GPT-5 nano, GPT-4o mini, GPT-OSS-20B, GPT-OSS-120B, Haiku~4.5, and Sonnet~4.6.
Replacement conditions use two LLM calls per night, one artifact call and one final writer call; \systemname{} and prompting baselines use one call.
We therefore report end-to-end cost and latency per insight.

\section{Evaluation}
\label{sec:eval}

All metrics are computed deterministically from saved per-night JSON traces.
\metric{SchemaErr} is the final JSON parse/schema error rate.
\metric{NumErr} is the fraction of extracted numeric claims that are unsupported by, or inconsistent with, the source fact bank.
Outputs with zero extracted numeric claims are excluded from \metric{NumErr}, so denominators vary across conditions.
\metric{SelErr} is the fraction of outputs whose selected metric contradicts the selection rule given to both the deterministic ranker and the LLM prompt.
\metric{AttrErr} is the fraction of outputs that make an attribution not supported by the evidence-gated input.

\begin{figure*}[t]
\centering
\includegraphics[width=0.85\textwidth]{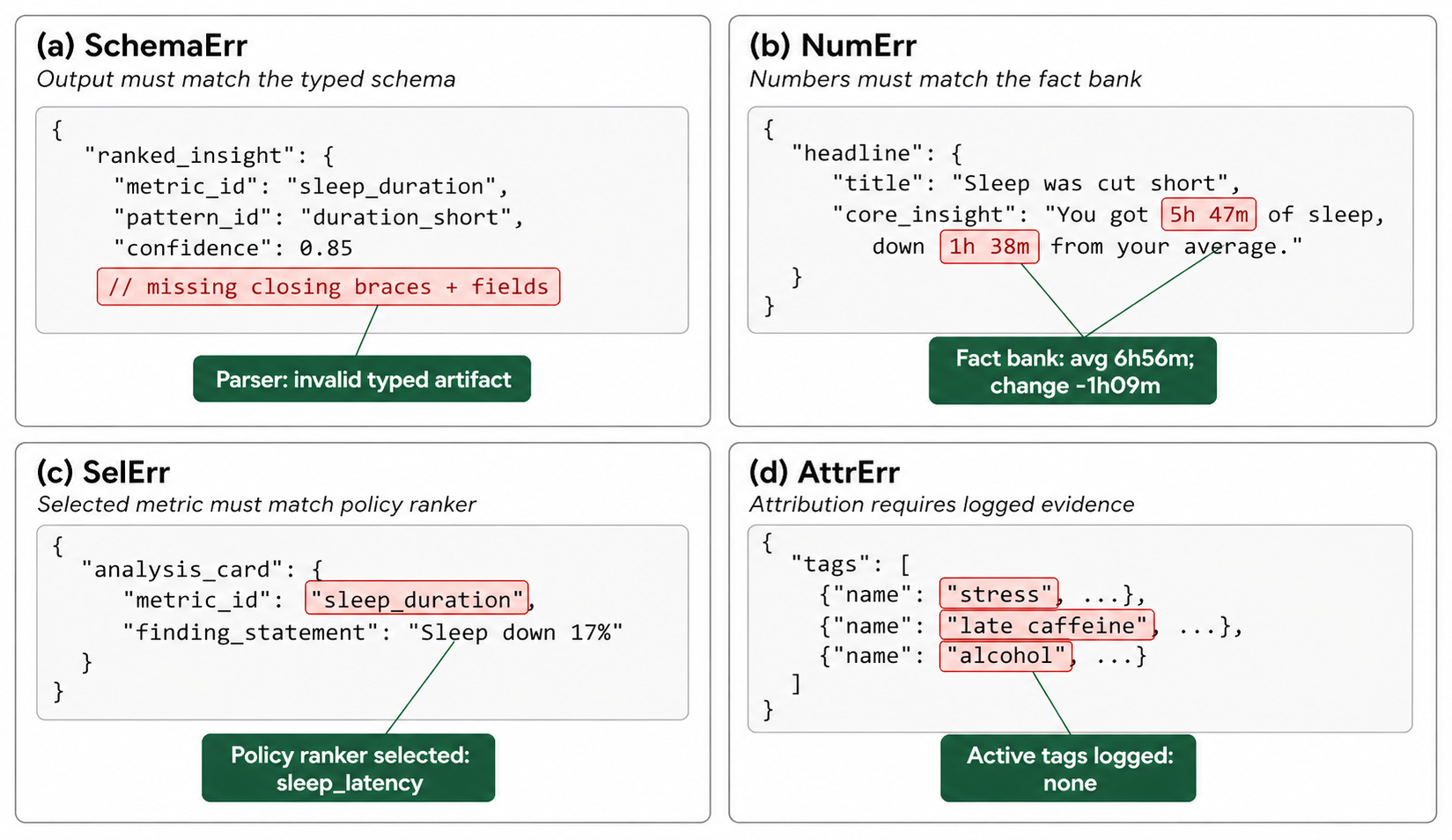}
\caption{Annotated examples of error types. Red highlights mark the LLM output span that violates the target rule; green callouts show the deterministic reference used by the evaluator. \metric{SchemaErr} captures parse/schema failures, \metric{NumErr} captures numeric hallucinations, \metric{SelErr} captures selection-policy mismatch, and \metric{AttrErr} captures unsupported attribution.}
\label{fig:errors}
\end{figure*}

\section{Results}
\label{sec:results}

\paragraph{Error--cost frontier.}
\Cref{fig:frontier} shows the main comparison.
Across all six models, \systemname{} occupies the low-cost, low-error region: average \metric{NumErr} is 0.7\%, the worst model remains at 2.0\%, and instruction-compliance error stays below 3\%.
The one-call baselines do not recover this frontier.
\condition{Structured Zero-Shot} often selects the wrong metric and makes unsupported numeric claims; \condition{Structured Few-Shot} improves schema and attribution behavior, but aggregate \metric{SelErr} remains 68.6\% (see \cref{tab:full_ablation}) and numeric errors remain high for several models.
This should not be read as a negative result for prompting: prompting helps formatting and some safety language, but it does not reliably reproduce the recurring analytical decisions that drive the insight.

\paragraph{Layer replacements isolate failure modes.}
\Cref{tab:full_ablation} complements the frontier by locating which analytical responsibility fails when delegated to an LLM.
\condition{Replace Comparison} delegates numerical observations and baseline comparisons to an LLM artifact; the aggregate \metric{NumErr} rises to 16.9\%.
\condition{Replace Ranker} keeps numeric facts grounded but degrades adherence to selection rules, with an aggregate \metric{SelErr} of 65.8\%.
\condition{Replace Attribution} preserves selection and numeric grounding but increases attribution errors, with aggregate \metric{AttrErr} of 24.5\%.
The pattern is more informative than a single end-to-end score: arithmetic, ranking policy, and evidence-gated attribution fail in different ways, and each is easier to verify as deterministic code than as prompted generation.

\paragraph{The writer interface matters.}
\condition{Replace Handoff} tests a more subtle boundary.
Upstream computation, ranking, and attribution remain deterministic, yet replacing the compact evidence-to-writer handoff raises aggregate \metric{NumErr} to 4.8\% and \metric{AttrErr} to 10.7\%.
This suggests that the interface between verified facts and prose is part of the reliability mechanism, not just prompt formatting.
A final writer can be useful, but the evidence packet that constrains what it may say should remain deterministic.

\paragraph{Representative examples clarify the metrics.}
\Cref{fig:errors} shows representative examples for the four main error families.
The examples are not used as evidence by themselves; they make the deterministic metrics in \cref{tab:full_ablation} interpretable.
They also show why schema-valid output is not enough: a response can parse correctly while still using an unsupported number, selecting the wrong metric, or adding attribution with no evidence above threshold.

\section{Discussion and Limitations}
\label{sec:discussion}

For recurring structured health outputs, deterministic code should own stable, verifiable responsibilities: arithmetic over records, selection policy, evidence-gated attribution, and the writer interface that controls which facts may enter prose.
LLMs remain useful where their strengths match the task: converting verified facts into natural, empathetic, schema-constrained language.
This framing complements RAG, tool use, and LLM-pipeline optimization \citep{lewis2020retrieval,schick2023toolformer,yao2023react,mialon2023augmented}: those methods can improve learned calls or retrieve better context, while our question is which repeated health-generation responsibilities should be removed from runtime prompting entirely.
Another promising direction is build-time agentic AI, where LLM agents help humans construct and audit deterministic health-insight pipelines before deployment, closer to pipeline-compilation systems and wearable-health agent case studies than to the runtime prompting baselines evaluated here \citep{khattab2023dspy,merrill2026phia}.

The partitioning principle should transfer most directly to domains with structured inputs, stable analytical rules, repeated deployment, and verifiable intermediates.
Lab result explanations, vitals monitoring summaries, medication-adherence coaching, and chronic-disease reports often share this shape.
We have not validated those domains here, and tasks requiring open-ended clinical reasoning, diagnosis, or treatment planning may need a different partition.

This study evaluates preservation of a deterministic reference system, not clinical optimality.
\metric{SelErr} is disagreement with an application selection rule rather than ground-truth health importance, and \metric{AttrErr} is attribution-policy compliance rather than biological causality.
Human evaluation of helpfulness and perceived personalization is complementary future work.

\section{Conclusion}
\label{sec:conclusion}

Structured health generation requires both reliable analytical computation and natural user-facing expression.
In our sleep-health case study, one-call prompting cannot match a deterministic analytical pipeline plus bounded writer, even with few-shot examples.
Layer replacement shows why: numerical comparison, ranking policy, attribution gates, and writer handoff each encode responsibilities that schema-valid LLM artifacts often fail to preserve.
\systemname{} supports a practical design rule for structured health systems with recurring analytical work: let code think fast over stable facts and policies, then let the LLM talk smart within verified bounds.

\bibliography{sd4h_references,references}
\bibliographystyle{icml2026}

\newpage
\appendix
\onecolumn

\section*{Appendix}
\section{Full Results}

\begin{table*}[h]
\caption{Full condition-by-model results. Values are percentages except $n$, \metric{Cost/night}, and latency. \metric{SchemaErr} is the final parse/schema failure rate; \metric{SelErr} is disagreement with the selection rule; \metric{AttrErr} is unsupported attribution. Replacement conditions include both the artifact call and final writer call in \metric{Cost/night}.}
\label{tab:full_ablation}
\vskip 0.05in
\centering
\tiny
\setlength{\tabcolsep}{2.6pt}
\begin{tabular}{llrrrrrrr}
\toprule
Condition & Model & $n$ & SchemaErr & NumErr & SelErr & AttrErr & Cost/night & Lat. \\
\midrule
\multirow{6}{*}{\systemname{}} & GPT-5 nano & 280 & 0.0 & 0.4 & 0.0 & 2.5 & \$0.0005 & 4.4s \\
 & GPT-4o mini & 280 & 0.0 & 0.4 & 0.0 & 0.0 & \$0.0007 & 8.1s \\
 & GPT-OSS-20B & 280 & 0.0 & 0.4 & 0.0 & 0.0 & \$0.0009 & 2.6s \\
 & GPT-OSS-120B & 280 & 0.0 & 2.0 & 0.0 & 0.4 & \$0.0023 & 4.5s \\
 & Haiku 4.5 & 280 & 0.0 & 0.3 & 0.0 & 0.0 & \$0.0063 & 8.6s \\
 & Sonnet 4.6 & 280 & 0.0 & 0.4 & 0.0 & 0.0 & \$0.0218 & 15.1s \\
\midrule
\multirow{6}{*}{Structured Zero-Shot} & GPT-5 nano & 280 & 0.0 & 54.2 & 95.0 & 18.6 & \$0.0010 & 4.4s \\
 & GPT-4o mini & 280 & 0.0 & 20.7 & 93.2 & 42.2 & \$0.0014 & 8.6s \\
 & GPT-OSS-20B & 280 & 0.4 & 15.5 & 71.8 & 2.5 & \$0.0020 & 4.8s \\
 & GPT-OSS-120B & 280 & 0.0 & 21.5 & 61.1 & 2.5 & \$0.0053 & 7.5s \\
 & Haiku 4.5 & 280 & 20.7 & 11.0 & 77.5 & 0.4 & \$0.0157 & 12.0s \\
 & Sonnet 4.6 & 280 & 32.5 & 24.2 & 82.5 & 1.1 & \$0.0538 & 16.8s \\
\midrule
\multirow{6}{*}{Structured Few-Shot} & GPT-5 nano & 280 & 0.0 & 3.4 & 71.1 & 3.6 & \$0.0004 & 3.8s \\
 & GPT-4o mini & 280 & 0.0 & 21.9 & 70.4 & 4.6 & \$0.0005 & 8.7s \\
 & GPT-OSS-20B & 280 & 0.0 & 38.9 & 68.9 & 2.1 & \$0.0008 & 2.4s \\
 & GPT-OSS-120B & 280 & 0.0 & 37.0 & 71.1 & 0.4 & \$0.0020 & 3.9s \\
 & Haiku 4.5 & 280 & 0.0 & 26.5 & 67.1 & 0.7 & \$0.0055 & 6.3s \\
 & Sonnet 4.6 & 280 & 0.0 & 14.9 & 63.2 & 0.7 & \$0.0189 & 9.9s \\
\midrule
\multirow{6}{*}{Replace Reference Report} & GPT-5 nano & 280 & 0.0 & 0.6 & 8.9 & 11.8 & \$0.0009 & 8.2s \\
 & GPT-4o mini & 280 & 0.0 & 0.4 & 4.6 & 11.4 & \$0.0012 & 14.5s \\
 & GPT-OSS-20B & 280 & 0.4 & 1.8 & 1.4 & 10.0 & \$0.0024 & 7.5s \\
 & GPT-OSS-120B & 280 & 0.0 & 3.4 & 0.7 & 10.7 & \$0.0060 & 12.9s \\
 & Haiku 4.5 & 280 & 0.7 & 3.0 & 2.1 & 8.9 & \$0.0115 & 15.4s \\
 & Sonnet 4.6 & 280 & 0.0 & 1.6 & 3.2 & 10.0 & \$0.0381 & 24.0s \\
\midrule
\multirow{6}{*}{Replace Comparison} & GPT-5 nano & 280 & 0.0 & 21.1 & 76.8 & 1.8 & \$0.0014 & 14.8s \\
 & GPT-4o mini & 280 & 0.0 & 20.9 & 68.2 & 5.7 & \$0.0021 & 31.0s \\
 & GPT-OSS-20B & 280 & 1.4 & 6.6 & 25.4 & 8.6 & \$0.0031 & 9.3s \\
 & GPT-OSS-120B & 280 & 0.0 & 8.8 & 15.0 & 10.0 & \$0.0077 & 16.3s \\
 & Haiku 4.5 & 280 & 0.0 & 26.7 & 34.6 & 8.6 & \$0.0225 & 23.7s \\
 & Sonnet 4.6 & 280 & 0.0 & 17.6 & 27.5 & 9.3 & \$0.0632 & 34.6s \\
\midrule
\multirow{6}{*}{Replace Ranker} & GPT-5 nano & 280 & 0.7 & 0.3 & 63.9 & 8.9 & \$0.0014 & 6.9s \\
 & GPT-4o mini & 280 & 0.0 & 0.0 & 71.8 & 9.6 & \$0.0022 & 13.5s \\
 & GPT-OSS-20B & 280 & 0.4 & 0.5 & 62.5 & 6.8 & \$0.0025 & 6.0s \\
 & GPT-OSS-120B & 280 & 0.0 & 3.1 & 67.1 & 10.7 & \$0.0070 & 10.2s \\
 & Haiku 4.5 & 280 & 5.3 & 2.2 & 62.9 & 7.5 & \$0.0160 & 11.9s \\
 & Sonnet 4.6 & 280 & 17.8 & 1.0 & 66.4 & 8.6 & \$0.0499 & 17.6s \\
\midrule
\multirow{6}{*}{Replace Attribution} & GPT-5 nano & 280 & 0.0 & 0.3 & 9.6 & 10.3 & \$0.0015 & 9.0s \\
 & GPT-4o mini & 280 & 0.0 & 0.4 & 2.9 & 21.1 & \$0.0020 & 12.7s \\
 & GPT-OSS-20B & 280 & 0.0 & 0.8 & 0.0 & 17.1 & \$0.0023 & 5.5s \\
 & GPT-OSS-120B & 280 & 0.7 & 5.0 & 0.7 & 22.1 & \$0.0058 & 8.0s \\
 & Haiku 4.5 & 280 & 0.4 & 1.1 & 1.1 & 32.8 & \$0.0198 & 22.9s \\
 & Sonnet 4.6 & 280 & 0.4 & 1.3 & 0.4 & 43.6 & \$0.0610 & 28.7s \\
\midrule
\multirow{6}{*}{Replace Handoff} & GPT-5 nano & 280 & 0.0 & 7.9 & 2.9 & 16.8 & \$0.0010 & 8.9s \\
 & GPT-4o mini & 280 & 0.0 & 1.5 & 2.9 & 11.4 & \$0.0014 & 14.2s \\
 & GPT-OSS-20B & 280 & 0.0 & 1.9 & 0.4 & 10.0 & \$0.0020 & 5.6s \\
 & GPT-OSS-120B & 280 & 0.0 & 3.8 & 0.4 & 9.3 & \$0.0046 & 8.1s \\
 & Haiku 4.5 & 280 & 0.4 & 8.1 & 0.4 & 8.6 & \$0.0133 & 17.9s \\
 & Sonnet 4.6 & 280 & 0.0 & 5.8 & 0.4 & 7.8 & \$0.0423 & 27.0s \\
\bottomrule
\end{tabular}
\end{table*}
\newpage
\section{Example Pipeline Trace}

\begin{figure*}[h]
\centering
\setlength{\fboxsep}{5pt}
\begin{tabular}{@{}p{0.48\textwidth}@{\hspace{0.03\textwidth}}p{0.48\textwidth}@{}}
\textbf{Input record and deterministic facts} &
\textbf{Bounded writer output} \\
\fcolorbox{black!20}{black!3}{\begin{minipage}[t]{0.45\textwidth}
\ttfamily\scriptsize
\{\\
\ \ "date": "2026-02-23",\\
\ \ "history\_window": "2026-02-09..2026-02-23",\\
\ \ "sleep": \{\\
\ \ \ \ "score": 84,\\
\ \ \ \ "duration": "7h 50m",\\
\ \ \ \ "deep": "1h 26m", "rem": "1h 40m",\\
\ \ \ \ "light": "4h 44m"\\
\ \ \},\\
\ \ "vitals": \{\\
\ \ \ \ "hrv": "34.2 ms",\\
\ \ \ \ "heart\_rate": "59.2 bpm",\\
\ \ \ \ "resp\_rate": "15.8 brpm"\\
\ \ \},\\
\ \ "snore\_percent": "6.0\%",\\
\ \ \textcolor{blue!60!black}{"tag\_candidates": [}\\
\ \ \ \ \textcolor{blue!60!black}{"Alcohol", "Stress", "Sick", "Fever"}\\
\ \ \textcolor{blue!60!black}{]},\\
\ \ "selected\_metric": "hrv",\\
\ \ "comparison": "34.2 vs 41.3 ms (-17\%)"\\
\}
\end{minipage}} &
\fcolorbox{black!20}{black!3}{\begin{minipage}[t]{0.45\textwidth}
\ttfamily\scriptsize
\{\\
\ \ "headline": \{\\
\ \ \ \ "title": "Your recovery looks different",\\
\ \ \ \ "core\_insight": "Your HRV dropped to\\
\ \ \ \ \ \ 34 ms, down 17\% from your baseline.",\\
\ \ \ \ "how\_to\_improve": "Do a 10-minute\\
\ \ \ \ \ \ progressive muscle relaxation tonight..."\\
\ \ \},\\
\ \ "analysis\_card": \{\\
\ \ \ \ "metric\_id": "hrv",\\
\ \ \ \ "finding\_statement": "HRV down 17\%",\\
\ \ \ \ \textcolor{blue!60!black}{"tags": [}\\
\ \ \ \ \ \ \textcolor{blue!60!black}{\{"name": "Alcohol", ...\},}\\
\ \ \ \ \ \ \textcolor{blue!60!black}{\{"name": "Stress", ...\},}\\
\ \ \ \ \ \ \textcolor{blue!60!black}{\{"name": "Sick", ...\},}\\
\ \ \ \ \ \ \textcolor{blue!60!black}{\{"name": "Fever", ...\}}\\
\ \ \ \ \textcolor{blue!60!black}{]}\\
\ \ \}\\
\}
\end{minipage}}
\end{tabular}

\vspace{0.08in}
\fcolorbox{green!40!black}{green!5}{\begin{minipage}{0.94\textwidth}
\small
\textbf{Deterministic handoff.} The pipeline selects \texttt{hrv} as the surfaced metric and gives the writer the fixed comparison ``34.2 vs 41.3 ms ($-17\%$).'' The writer may copy the pre-generated attribution tags shown in blue, but it may not add new tags or recalculate numbers.
\end{minipage}}
\caption{Representative \systemname{} trace for one user-night (Sonnet 4.6, 2026-02-23). The figure shows the compacted structured input, deterministic selected facts, and final schema-constrained output. Blue spans mark tag fields copied through the bounded writer interface.}
\label{fig:example_trace}
\end{figure*}

\end{document}